\documentclass[letterpaper]{article} 
\usepackage{aaai2027}  
\usepackage[hyphens]{url}  
\usepackage{graphicx} 
\usepackage{natbib}  
\usepackage{caption} 
\usepackage{algorithm}
\usepackage{algorithmic}
\usepackage{booktabs}
\usepackage{amsmath}
\usepackage{amssymb}
\usepackage{multirow}
\usepackage{array}
\usepackage{colortbl}

\definecolor{cLightBlue}{RGB}{217,231,244}
\definecolor{cLightPink}{RGB}{255,199,206}
\definecolor{cLightYel}{RGB}{255,249,217}
\definecolor{cLightGrn}{RGB}{188,226,195}
\newlength{\dltwd}
\newcommand{\up}[1]{\makebox[\dltwd][l]{\textsubscript{\textcolor{green!55!black}{#1}}}}
\newcommand{\dn}[1]{\makebox[\dltwd][l]{\textsubscript{\textcolor{red!75!black}{#1}}}}
\newcommand{\nd}{\makebox[\dltwd][l]{}}
\newcommand{\best}[1]{\textbf{#1}}
\newcommand{\snd}[1]{\underline{#1}}

\title{GoAnt: Quality-Diversity Multi-Agent Search for Alpha Factor Discovery in Market Microstructure Data}
\author{
    Stella Zhao\textsuperscript{\rm 1},
    Tommy Sha\textsuperscript{\rm 2}
}
\affiliations{
    \textsuperscript{\rm 1}University of Minnesota, Minneapolis, MN, USA\\
    \textsuperscript{\rm 2}Stony Brook University, Stony Brook, NY, USA\\
    zhao2052@umn.edu, tianming.sha@stonybrook.edu
}
\begin{document}

\maketitle

\begin{abstract}
Automated alpha factor discovery searches symbolic trading signals from
price--volume panels and order-book data under a fixed evaluation budget.
Existing single- and multi-agent program-search systems can overfit predictive
proxies that fail after execution costs and repeatedly explore redundant factor
families, limiting execution robustness and behavioral diversity. We introduce
GoAnt, a quality-diversity multi-agent search framework that combines
non-communicating Explorer, Exploiter and Connector workers with a shared
adaptive Mental Map and a compact Queen dispatcher. The Mental Map organizes
candidates by leakage-free execution profiles and retains one elite per niche,
while the Queen reallocates the evaluation budget from explicit search-state
summaries. We also define a map-independent effective-yield protocol that
counts high-quality, mutually nonredundant factors directly from each method's
evaluation records, giving archive-based and map-free systems the same ruler.
On real A-share microstructure data spanning 2023--2026, GoAnt reaches
quality-weighted yields of 41.8 and 47.6 in price--volume and order-book
settings, improving the strongest baseline by 57\% and 97\% under matched
budgets. Its locked populations retain 0.64 and 0.67 of in-sample quality out
of sample, compared with 0.61 and 0.63 for a static map.
\end{abstract}

\section{Introduction}

The automated discovery of alpha factors, mathematical signals predictive of asset returns, has advanced rapidly with the rise of Large Language Models (LLMs). A growing body of work now casts factor mining as a program synthesis task, in which an agent iteratively proposes, evaluates, and refines symbolic expressions. Despite this progress, moving from daily-frequency price--volume panels to intraday order-book data remains an open challenge, and the two categories are usually studied apart even though a coordination mechanism worth reporting should hold across both. 

As data granularity increases, agents confront the Execution Trap: factors can exhibit strong statistical predictive power (e.g., high RankIC) yet deliver negative net returns once hidden execution costs, slippage, and market impact \cite{almgren2001optimal} are taken into account. Single-agent setups that greedily optimize such naive metrics tend to collapse into localized sub-optima \cite{alphagpt2023}. 

To mitigate single-agent mode collapse, recent work has pivoted toward multi-agent collaboration, predominantly built on static topologies (e.g., a ``Pod Shop'' pairing an LLM Manager with LLM Analysts) \cite{metagpt2023, chatdev2023}. While effective for general-purpose reasoning, such systems encounter the Reasoning Trap of \citet{nju_orchestrator} in this setting: across highly noisy and vast financial search spaces, general-purpose off-the-shelf LLM coordinators are prone to consensus bias and context squeezing. As a result, they allocate computational budget poorly and yield homogenized factor populations.

To address these failures, we introduce \textbf{GoAnt}, a dynamic, decoupled multi-agent architecture with three components. Generation is offloaded to non-communicating Worker Ants organized as three parallel islands (Explorer, Exploiter, Connector), which together form an embodied Mixture-of-Experts \cite{shazeer2017outrageously}, that populate a single shared Mental Map, a Quality-Diversity (QD) archive (MAP-Elites) whose cell centres are fixed at birth and whose cardinality grows online as the population reaches new regions. Budget allocation is delegated to a compact Queen Ant: a Qwen2.5-Instruct \cite{qwen25} model LoRA-adapted by knowledge distillation \cite{hinton2015distilling} from eight frontier teachers, conditioned on explicit search-state metrics (per-island coverage, isolation, stagnation, and duplication). Exp~5 compares it against round-robin, EMA and UCB schedulers under an identical Atlas and budget.

Our central claim concerns the archive, not the orchestrator. The Atlas is dynamic in exactly one respect, that of adaptive capacity: a candidate lying outside every existing niche radius founds a new cell, so the archive's cardinality grows with the reachable region of behaviour space, while every cell centre stays fixed at the descriptor that founded it. Centre motion is run as a separate arm rather than bundled in, so that cardinality growth and centre mobility are attributed independently. Because growing an archive spends budget on exploration, we quantify what that resolution costs relative to a frozen map and report the result in both data categories.

The core contributions of this paper are:
\begin{itemize}
    \item We identify the Execution Trap and the Reasoning Trap that jointly limit current LLM-driven alpha mining, and show that neither is an artifact of a single data category, since both are exhibited in two.
    \item We propose GoAnt, a dynamic multi-agent system coupling an adaptive-capacity MAP-Elites Mental Map with decoupled island generators and a multi-teacher distilled orchestrator (the Queen Ant).
    \item We separate the two mechanisms that the term ``dynamic archive'' normally conflates, namely centre mobility and adaptive cardinality, and attribute them independently through a $2{\times}2$ factorial instead of reporting their sum.
    \item We define the static control by end-of-run invariants that abort a run on any centre move, cell creation or re-fit, so that the comparison measures what its label claims; and we hold the search operator fixed across geometry arms, so that a geometry contrast is not confounded by a change in how candidates are proposed.
    \item We introduce a map-independent effective-yield protocol that extracts a quality-filtered, pairwise nonredundant factor set from the common evaluation record, so archive-based and map-free systems are scored by the same ruler (Table~\ref{tab:exp1_main}; Supplementary Appendix~A).
\end{itemize}

\section{Related Work}
\subsection{LLM-Driven Program Search and Alpha Mining}
LLMs have been coupled with evaluator-in-the-loop program search and QD archives over code \cite{funsearch, elm, llmatic}, and formulaic-alpha discovery automated by reinforcement \cite{alphagen}, evolutionary \cite{autoalpha, alphaevolvefin} and LLM-guided \cite{alphagpt2023, alphajungle, alphaagent, factorminer} search. All work at a single granularity and none prices execution cost \cite{almgren2001optimal} inside the loop, so the cost wall that dominates at finer granularities never enters the objective.

\subsection{Fixed and Adaptive Search Structure}
Collaborative LLM frameworks span non-communicating sampling \cite{selfconsistency}, shared-memory societies \cite{genagents}, debate ensembles \cite{debate, chateval}, role graphs \cite{metagpt2023, chatdev2023} and quantitative-research pipelines \cite{rdagentq, tradingagents}. They vary the communication topology but all fix the partition of the solution space and the allocation policy in advance. GoAnt fixes its roles and generators and adapts that structure instead, which no prior system measures under a matched budget.

\subsection{Quality-Diversity and Archive Structure}
QD algorithms keep an archive of high performers spread over a behavioral space \cite{mouret2015illuminating, qdframework}, and its geometry is itself a design variable: CVT-MAP-Elites fixes a Voronoi partition \cite{cvtmapelites}, CMA-ME adapts the emitter but not the partition \cite{cmame}, sliding-boundary MAP-Elites moves boundaries with the evolved distribution \cite{mesb}, and dominated novelty search drops the archive altogether \cite{dns}. The Mental Map partitions by leakage-free execution profile, with centres fixed at birth like a CVT and cardinality growing online; we ablate centre mobility and adaptive capacity separately under matched budgets in two data categories. We report behavioral coverage, not cross-sectional correlation.

\begin{figure*}[t]
    \centering
    \includegraphics[width=1.0\textwidth]{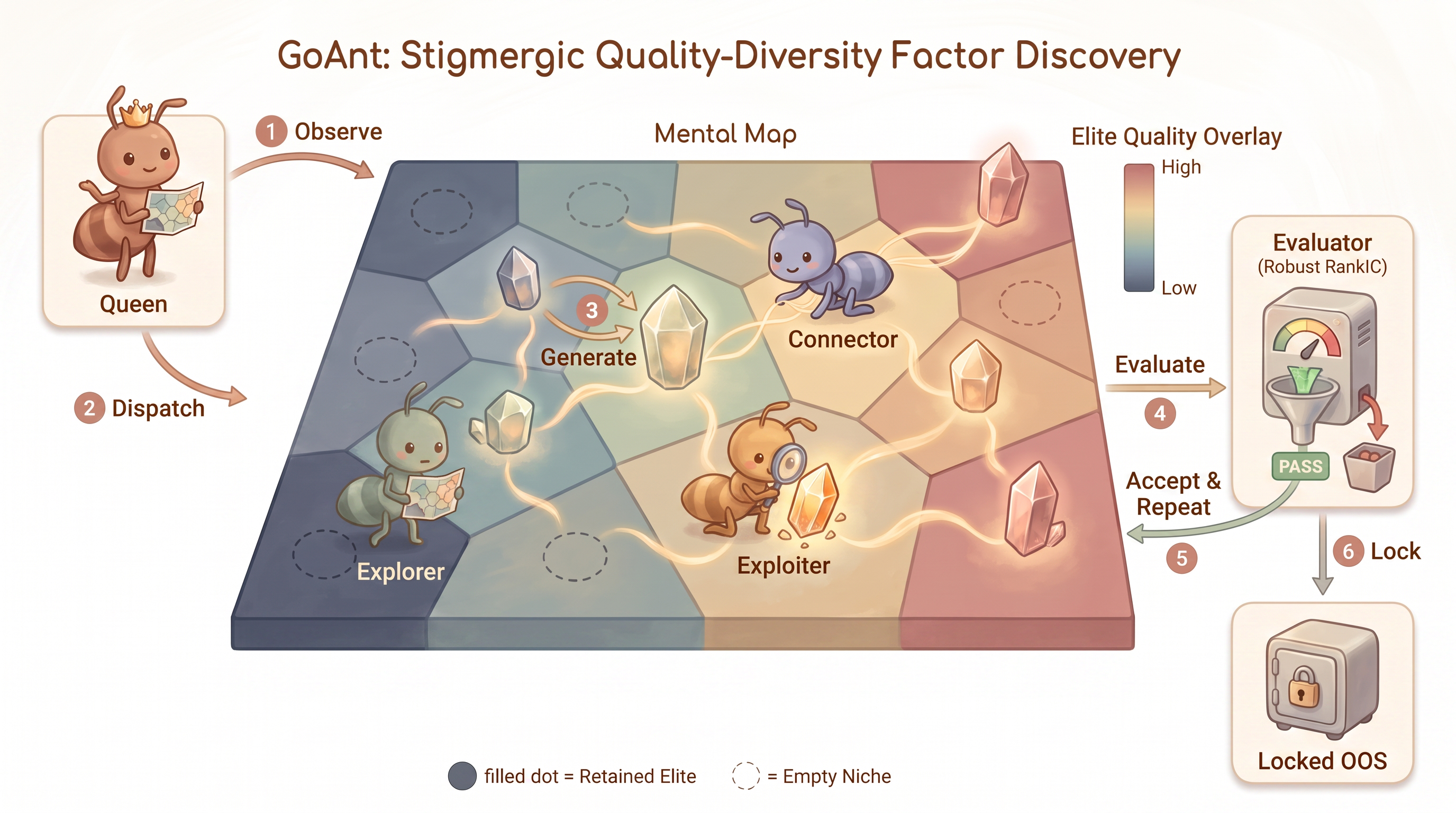}
    \caption{The GoAnt architecture. The distilled \textbf{Queen Ant} observes the \textbf{Mental Map} (\textbf{1}), a quality--diversity archive acting as shared stigmergic memory, and dispatches (\textbf{2}) one Explorer, Exploiter or Connector ant, which generates candidate factor logic (\textbf{3}). The \textbf{Evaluator} computes the deflated fitness and applies turnover and stability gates (\textbf{4}); a candidate that passes and improves on its niche's record is written back (\textbf{5}), and the surviving elites are locked for out-of-sample replay (\textbf{6}). The Map is drawn schematically: its cells live in the six-dimensional descriptor space, so plotted positions indicate adjacency only and no panel distance is a distance in $\mathcal{Z}$.}
\label{fig:architecture}
\end{figure*}

\section{Methodology: The GoAnt Framework}

We frame automated factor mining as budget-constrained quality--diversity (QD) search: under a fixed evaluation budget $B$, decoupled generators must fill an archive that is simultaneously high-quality and behaviorally diverse, without collapsing onto a single high-scoring program. GoAnt separates this problem into three components (Fig.~\ref{fig:architecture}): a niche-structured archive (the Mental Map), non-communicating generators (Worker Ants), and a distilled budget-allocation policy (the Queen Ant). We keep the treatment at the level of the algorithm: we prove the archive's update-rule invariants, and the allocation objective is a formulation whose effect we measure directly in the experiments.

A Queen Ant orchestrator allocates the budget across three parallel Worker-Ant islands (Explorer, Exploiter, Connector); each island generates candidates independently and all candidates compete for niches in one shared MAP-Elites archive. Worker roles stay fixed and message-free, and only the archive and the allocation over generators are adaptive.

\subsection{The Execution Trap and Formulation}
Let the search space of symbolic alpha factors be $\mathcal{F}$. Traditional systems maximize a predictive proxy $\max_{f \in \mathcal{F}} \mathbb{E}[R(f)]$ (e.g., RankIC), whereas the quantity of interest is the execution-aware net return $N(f, C)$ under cost $C$. The Execution Trap is that the two diverge once execution is priced: under a conservative full-spread cost wall, factors with strong RankIC yield negative $N$. Systems that select on the predictive proxy alone never meet this wall during search, so arms 1--3h are scored here under the same execution-aware evaluator and feasibility gates as GoAnt rather than on the RankIC they were originally tuned for. The divergence is mild on daily price--volume panels and severe on intraday order-book data, which is why we run both categories rather than only the one where it is most dramatic. GoAnt therefore adopts a layered objective: (i) the search fitness is a deflation-corrected signal-quality score $S(f)$ crediting a robust predictive $t$-statistic modulated by daily-IC stability, so noise whose daily IC flips sign is driven toward zero; (ii) turnover is offloaded to the descriptor axes of the Mental Map (below); and (iii) net return $N$ is retained for post-hoc validation and OOS gating only, never as a search target. Arm~7 tests that last choice directly in both categories.

\subsection{The Mental Map: A Feature-Space Quality-Diversity Archive}
To prevent mode collapse and factor crowding, the Mental Map is a Quality-Diversity archive over a behavioral feature space $\mathcal{Z}$, where a descriptor operator $\phi: \mathcal{F} \to \mathcal{Z}$ embeds each evaluated factor by its label-free execution profile. The admission rule for an axis is constructive and statically auditable: a quantity may enter $\phi$ only if it is computable without reading any future-return label, so no axis can be an argument of the search fitness. Six axes satisfy it: turnover and mean effective spread describe execution, operator-chain depth describes structure, and signal autocorrelation, cross-sectional skewness and long/short spread tilt describe the shape of the position the factor actually takes. Each is robustly standardized and clipped at $\pm 3\sigma$, with the scaling statistics fitted once per data category on the development split so no look-ahead enters through the scaler. The same six axes serve both categories; only the scaler differs, and every contrast is between arms within one category. A niche is a region of $\mathcal{Z}$ retaining the elite discovered within it, ranked by signal quality $S(f)$ and not by raw net return \cite{mouret2015illuminating}. The Map is dynamic in exactly one respect. Instead of a fixed lattice, GoAnt maintains a growing set of niche nodes joined by mutual $k$-nearest-neighbor edges: a candidate lying farther than one niche radius from every existing node founds a new node at its own descriptor, and that centre never moves again. The tessellation gains resolution over a run but is never redrawn, and no centre is ever dragged by the data assigned to it. Freezing the node set as well recovers a fixed nearest-site partition, a grid or a CVT tessellation \cite{cvtmapelites}, which is the control we compare against. Table~\ref{tab:dynamism} separates whether growing the map pays from centre motion at a single matched budget, identifying the mechanism responsible for a difference without establishing how the answer would move with the budget.

\medskip\noindent\textbf{Archive invariants, and what we actually score.} Let $A_t$ be the archive after $t$ evaluations, $\mathcal{O}_t$ its set of occupied niches, and $e_{k,t}$ the elite of niche $k$. The update rule admits a candidate only if it is the highest-scoring factor seen so far within its own niche, so $|A_t|=|\mathcal{O}_t|$.

\noindent\textbf{Proposition (Elitist monotonicity).} A niche's elite is replaced only by a strictly higher-scoring factor, so $S(e_{k,t+1})\ge S(e_{k,t})$ for every niche occupied at both updates. Because centres are fixed at birth this holds for the entire run and not only between re-charting events: founding a new node can divert future candidates from an existing niche, but never moves a centre and never lowers a stored elite. The periodically re-fitted arm (6b) forfeits the guarantee by construction, and we report its realized best-so-far trajectory empirically instead.

\noindent\textbf{Observation (coverage identity).} With one elite per occupied niche, uniformly sampling an archived elite makes its niche index uniform on $\mathcal{O}_t$, so the entropy of the sampled elite is exactly $\log|\mathcal{O}_t|$ \cite{shannon1948mathematical}. Because occupancy is the only quantity this archive can widen, this entropy restates the cell count and the identity constrains redundancy only within a cell, leaving correlation across cells unaddressed. Every QD and coverage figure we report is therefore computed on a fixed, method-independent grid of four factor families $\times$ five turnover bins, held identical across all arms and budgets, so no arm is scored on a partition it drew itself.

\subsection{Worker Ants: Non-Communicating Generation}
Unlike collaborative chatting agents, GoAnt's Worker Ants exchange no direct messages and coordinate only through the shared Map:
\begin{itemize}
    \item \textbf{Explorer Ants}: Perform radical, high-temperature structural mutations to break local optima.
    \item \textbf{Exploiter Ants}: Execute intensive, localized hill-climbing via Learned Repair on known elites.
    \item \textbf{Connector Ants}: Perform semantic crossover, combining genes from spatially distant niches on the Mental Map.
\end{itemize}

\subsection{The Queen Ant: A Distilled Orchestrator}
Under a fixed evaluation budget, the orchestration layer must repeatedly decide which Worker-Ant island should receive the next allocation. A natural baseline treats each island as an arm of a multi-armed bandit under an Upper Confidence Bound (UCB) rule \cite{auer2002finite, lai1985asymptotically}. On noisy financial rewards this is myopic: it follows instantaneous empirical rewards and, once a niche yields high marginal gains, hyper-concentrates on it until duplication dominates the proposal stream and the search stalls. GoAnt instead instantiates the orchestrator as a compact language model. This Queen Ant (Qwen2.5-Instruct \cite{qwen25}, LoRA-adapted \cite{hu2021lora}, at the 1.5B capacity evaluated in Exp~5) is trained on allocation decisions distilled from frontier teachers. Exp~5 measures its cost against a scheduler that needs no GPU at all.

We construct the training signal by knowledge distillation from an ensemble of eight frontier teacher models (GPT-5.6, Claude Opus 4.8, Claude Opus 5, DeepSeek-V4-Pro, Qwen3.7-Max, GLM-5.2, Kimi K2.7-Code, and MiniMax-M3). For each orchestration state, which summarizes per-island pull counts, empirical Q-values, coverage, isolation and stagnation, the teachers produce chain-of-thought rationales and select an arm over \{Explorer, Exploiter, Connector\}. The Queen Ant is then supervised (LoRA SFT on a 35M-token distilled corpus) to reproduce the selection and its structured justification, for instance an Explorer action where UCB would keep pulling a saturated niche. The resulting orchestrator conditions its arm choice on a natural-language summary of the global search state instead of on a scalar bandit statistic.

\medskip\noindent\textbf{Allocation objective.} The multi-objective reward above can be stated explicitly. Let $\pi$ be the allocation policy, $\tau_B$ the sequence of proposals produced once the full budget $B$ is spent, and $A_B$ the resulting archive. The Queen is distilled toward
\begin{equation}
\begin{split}
\max_{\pi}\ \; J(\pi)\;=\;&\alpha\,\bar{C}(A_B) \;+\; \beta\,\bar{Q}(A_B) \;-\; \gamma\,\bar{D}(\tau_B),\\
&\alpha,\beta,\gamma\ge 0,\quad \alpha+\beta+\gamma=1,
\end{split}
\end{equation}
where $\bar{C},\bar{Q},\bar{D}\in[0,1]$ are the coverage, quality and redundancy terms rescaled to a common range so that the weights form a convex combination: $C(A_B)$ counts covered niches, $Q(A_B)$ aggregates elite quality (e.g.\ the QD-score $\sum_k S(e_k)$), and $D(\tau_B)$ counts duplicated or invalid proposals (a property of the trajectory, not of the final archive).

\medskip\noindent\textbf{Coverage is earned, not issued.} Under adaptive capacity a candidate far from every existing centre founds a cell and occupies it in the same step, so counting occupied niches directly would reward the mere proposal of outliers. Coverage therefore admits a newly founded cell only once its elite clears the pre-registered quality floor $q^\star$ used for reporting, which makes each new cell a bet that has to be paid off:
\begin{equation}
C(A_B)\;=\;\bigl|\{\,k \;:\; \mathrm{niche}\ k\ \mathrm{occupied}\ \wedge\ S(e_k) > q^\star \,\}\bigr|.
\end{equation}
Cell creation therefore costs an evaluation but pays nothing until the cell is populated by a factor that would have counted anyway, which makes exploration a bet, not a subsidy and leaves the objective bounded by the same floor the results tables use. Exp~5 measures whether the distilled Queen reaches higher coverage and quality at lower redundancy than round-robin, UCB, and heuristic schedulers under an identical budget, and the out-of-sample value of the resulting diversity is tested in Exp~4.

\section{Experiments}\label{sec:exp}

Every number reported in this section is produced by a single driver under one
fixed configuration, so that the arms differ in exactly one respect at a time.
All arms share the same data and day split, the same skeleton and
feasibility gates, the same evaluator, the same search fitness $S$, and the same
evaluation budget; the only variable is the coordination and memory structure
under test. Every arm is run with five independent seeds and every table in the body reports the
mean over them. Because factor quality is heavy-tailed and one fortunate seed can
dominate a mean while leaving the median untouched, dispersion is quantified by a
95\% bootstrap confidence interval resampled over seeds rather than over the
factors within a run, and arms are compared with the rank-based procedure set out in
Appendix~G of the supplementary material.

\begin{table*}[tp]
\centering
\footnotesize
\setlength{\tabcolsep}{5.2pt}
\renewcommand{\arraystretch}{1.18}
\begin{tabular}{@{}llc rrrrr rrrrr@{}}
\toprule
& & & \multicolumn{5}{c}{\textbf{Price--Volume (PV)}} & \multicolumn{5}{c}{\textbf{Order-Book (L2)}} \\
\cmidrule(lr){4-8}\cmidrule(lr){9-13}
\# & Method & Topology & distinct & q-wtd\nd & best & $|\rho|@q^\star$ & cells & distinct & q-wtd\nd & best & $|\rho|@q^\star$ & cells \\
\midrule
\multicolumn{13}{@{}l}{\textbf{LLM-driven baselines, static coordination}} \\
1  & Single-Agent          & greedy   & 6  & 5.8\nd  & 5.12 & 0.68 & ---  & 5  & 4.6\nd  & 4.71 & 0.71 & ---  \\
2  & Debate/Vote           & static   & 15 & 16.3\nd & 5.81 & 0.52 & ---  & 13 & 13.9\nd & 5.34 & 0.56 & ---  \\
3a & RD-Agent-Quant        & static   & 21 & 22.4\nd & 6.24 & 0.35 & ---  & 19 & 20.1\nd & 5.88 & 0.38 & ---  \\
3b & TradingAgents         & static   & 19 & 19.6\nd & 5.93 & 0.37 & ---  & 17 & 17.4\nd & 5.52 & 0.40 & ---  \\
3c & AlphaAgent~\shortcite{alphaagent} & static & 23 & 25.1\nd & 6.41 & 0.39 & ---  & 20 & 21.6\nd & 6.02 & 0.41 & ---  \\
\midrule
\multicolumn{13}{@{}l}{\textbf{Non-LLM factor-search baselines}} \\
3d & AlphaGen~\shortcite{alphagen}     & n/a & 20 & 20.8\nd & 6.08 & 0.35 & ---  & 18 & 18.5\nd & 5.71 & 0.37 & ---  \\
3e & AutoAlpha (GP)~\shortcite{autoalpha} & n/a & 24 & 26.6\nd & 6.55 & 0.36 & ---  & 22 & 24.2\nd & 6.12 & 0.37 & ---  \\
\midrule
\multicolumn{13}{@{}l}{\textbf{Quality--diversity search baselines}} \\
3f & CVT-MAP-Elites~\shortcite{cvtmapelites} & fixed & 18 & 18.4\nd & 5.94 & 0.41 & 96 & 17 & 17.2\nd & 5.76 & 0.43 & 96 \\
3g & CMA-ME~\shortcite{cmame}          & fixed & 14 & 13.7\nd & 5.42 & 0.45 & 96 & 12 & 11.6\nd & 5.08 & 0.47 & 96 \\
3h & Random Search                     & none  & 7 & 6.5\nd & 4.63 & 0.44 & ---  & 6 & 5.5\nd & 4.35 & 0.46 & ---  \\
\midrule
\rowcolor{cLightGrn}
4  & \textbf{GoAnt (ours)} & \textbf{adaptive} & \best{34} & \best{41.8}\up{+57\%} & \best{7.62} & \best{0.31} & \best{118} & \best{37} & \best{47.6}\up{+97\%} & \best{7.84} & \best{0.28} & \best{131} \\
\midrule
\multicolumn{13}{@{}l}{\textbf{Ablations of the Mental Map}} \\
5  & No-Map                & none     & 9  & 8.7\nd  & 7.14 & 0.71 & ---  & 8  & 7.6\nd  & 7.29 & 0.74 & ---  \\
6  & Static Map            & frozen   & 27 & 30.5\nd & 7.21 & 0.38 & 96   & 26 & 31.2\nd & 7.32 & 0.39 & 96   \\
6b & Re-fit Atlas          & periodic & 29 & 33.7\nd & 7.06 & 0.36 & 104  & 29 & 34.8\nd & 7.22 & 0.36 & 109  \\
6c & Moving-Centre         & semi     & \snd{31} & \snd{36.2}\nd & \snd{7.34} & \snd{0.34} & 121  & \snd{31} & \snd{39.4}\nd & \snd{7.58} & \snd{0.33} & 134  \\
7  & Raw-Net obj.          & adaptive & 11 & 9.9\nd  & 4.87 & 0.63 & 115  & 8  & 6.8\nd  & 4.12 & 0.69 & 128  \\
\bottomrule
\end{tabular}
\caption{\textbf{Exp 1, the coordination spectrum under one ruler, reported independently on both data categories.} Arms 1--3c are static-topology LLM systems, 3d--3e non-LLM factor search, and 3f--3h standard quality--diversity or random search; all are reimplemented under our generator interface, evaluator and budget. Arms 1--3h hold their archive geometry fixed for the entire run; arm 4 grows archive cardinality online with centres frozen at birth; arms 5--7 each vary one mechanism of that design, with 6b and 6c reshaping the geometry in the two other ways a map can. \textbf{Bold} is the column best, \underline{underline} the runner-up, and the green subscript arm 4's margin over arm 3e, the strongest baseline. \texttt{distinct} counts factors above the pre-registered floor $q^\star$ with pairwise $|$Spearman$|$ below $\rho$; \texttt{q-wtd} weights them by their excess over $q^\star$; $|\rho|@q^\star$ is taken among those factors (lower is better); \texttt{cells} is the realized archive cardinality, for archive-based arms only.}
\label{tab:exp1_main}
\end{table*}

\subsection{Data}
\paragraph{Two data categories, one universe.} We run every arm on real A-share microstructure data spanning 2023--2026, instantiated as two distinct categories over a shared instrument universe: price--volume (PV) panels and order-book (L2) snapshots. Because they cover the same instruments over the same period, an arm receives identical coordination machinery, an identical budget and an identical evaluator in both, and the only thing that varies is how much microstructure the descriptor axes and the cost model can resolve. Every results table reports the two side by side, not pooled, so that each entry is the same claim measured under two views of the same universe.

\paragraph{Why a single category is not enough.} A coordination result obtained on one data source is not separable from an overfit to that source, and the two categories fail in opposite directions. Order-book data carries a noise structure that a niche-based memory could plausibly be exploiting, whereas on price--volume panels the cost wall is a second-order correction rather than the dominant term and the Execution Trap is correspondingly mild. Measuring both therefore identifies which effects are properties of the coordination structure and which are properties of the data, and an effect that appears in one category but reverses in the other is reported as such instead of averaged away. The admission gates and the search fitness are defined identically in the two categories. The descriptor uses the same six axes in both, with only its scaling statistics fitted per category; the cost model is likewise category-specific. Both are fixed in advance per category and never tuned per arm, and because every reported contrast is between arms within a category, neither difference is ever load-bearing for a comparison.

\paragraph{Splits are disjoint by construction.} Within each data category the days are partitioned once into a development window, used for search and for fitting every method's grid, centres and descriptor scaler, and a strictly later locked window used only for the out-of-sample replay of Exp 4. No arm sees a locked day during search, and no scaler or tessellation is ever estimated on one.

\subsection{Protocol}

\paragraph{One ruler for every arm.} Map-based and map-free systems have no common archive to be read out of, so every arm is scored from its evaluation record, not from whatever structure it happens to maintain: a candidate counts if and only if it clears the same pre-registered admission floor, and the diversity of what survives is measured on a fixed, method-independent grid that no arm can influence. Fitness is deflated identically for every arm before any comparison is made, redundancy is always conditioned on quality instead of averaged over the whole population, and the descriptor scaler is fitted once on the development split and then frozen for all arms and all seeds. Appendix~A of the supplementary material states each of these commitments precisely and gives the reasoning behind it.

\subsection{Exp 1: The Coordination Spectrum}
To pinpoint which architectural property drives the gains, we evaluate a spectrum of coordination mechanisms under an identical data split, evaluator, feasibility gates, evaluation budget, and search fitness $S$; the only variable is how agents coordinate and remember (Table~\ref{tab:exp1_main}). The spectrum ranges from a single agent, through increasingly communicative but structurally static multi-agent baselines re-implemented from the literature, namely pre-evaluation debate/voting \cite{debate, chateval} and three factor-mining and trading pipelines \cite{rdagentq, tradingagents, alphaagent}, through two searchers that replace the LLM generator entirely, reinforcement-learning alpha search \cite{alphagen} and genetic programming \cite{autoalpha}, and through three archive-side controls that keep our generator but swap the Queen for a standard quality--diversity or random sampler \cite{cvtmapelites, cmame}, to GoAnt's adaptive-capacity Atlas and finally the map ablations. Because every arm shares the same generator, evaluator, and budget, and all methods are scored from their own evaluation record stream and not from a self-selected elite pool, the comparison isolates the coordination mechanism itself, not implementation artifacts.

The table is read along two contrasts. Arms 1--3h versus 4 test the paper's central claim, that a Mental Map organizes otherwise homogeneous agents, with 3f--3h isolating what the Queen contributes over an off-the-shelf archive; arms 4 versus 5--7 form the internal ablation that attributes any gain to the map, to its dynamism specifically, and to the layered objective. GoAnt's own dynamism is adaptive capacity alone: its cell centres are fixed at birth and never move, and the only thing that changes over a run is how many cells there are. Arms 6, 6b and 6c therefore probe three genuinely different things a map can do about its geometry: 6 fixes centres, cell count and boundaries for the entire run, 6b redraws the whole map on a schedule, and 6c is the reverse ablation that adds mean-tracking centre motion back on top of GoAnt, which isolates centre mobility as a mechanism separate from cardinality growth.

\subsection{Exp 2: Mechanism Ablations --- Isolating Map, Dynamics, and Objective}
The macro comparison establishes that the Mental Map helps; it does not establish which property of it does. We therefore hold every other component fixed and vary one mechanism at a time.

\begin{itemize}
    \item \textbf{Is the map necessary?} Arm 5 replaces the Atlas with a flat greedy pool while retaining the identical generator, evaluator and budget.
    \item \textbf{Which part of the dynamism is necessary?} Arm 6 is a fully frozen map and arm 6b a periodically redrawn one; arm 6c runs in the opposite direction, restoring mean-tracking centre motion on top of GoAnt. Each fixes a different aspect of the geometry, so the three together bracket what ``static'' can mean.
    \item \textbf{Does the layered objective matter?} Arm 7 keeps the Atlas and the descriptor axes unchanged and swaps only the archive-admission fitness for the raw net return, which is dominated by cost-wall nonlinearity.
\end{itemize}

\paragraph{Defining the static control.} A nominally frozen archive can still drift through code paths that look unrelated to geometry, so we define arm 6 by end-of-run invariants rather than by configuration: its centres, its cell count and its boundaries are compared against their values at initialization, and the run is aborted if any of them has changed. Appendix~B of the supplementary material reports the audit and the invariants it checks.

\paragraph{Decomposing the dynamism.} A dynamic archive differs from a fixed tessellation along two independent axes, and bundling them makes the resulting claim uninterpretable. The first axis is whether cell centres track the running mean of their members; the second is whether a candidate outside every existing niche radius may instantiate a new cell. We run the full $2\times2$ factorial over these two switches with everything else held fixed (Table~\ref{tab:dynamism}), so that the contribution of centre mobility and the contribution of adaptive cardinality are attributed separately, not jointly. Note the direction of the comparison: because GoAnt keeps its centres fixed, adding centre motion is an ablation away from our method, not towards it. We also run periodic re-charting, the strongest form of a dynamic map, so that both the frozen and the fully redrawn geometries appear as controls.

\begin{table}[tb]
\centering
\footnotesize
\setlength{\dltwd}{1.42em}\setlength{\tabcolsep}{3.6pt}
\renewcommand{\arraystretch}{1.15}
\begin{tabular}{@{}llcrrrr@{}}
\toprule
& & & \multicolumn{2}{c}{PV} & \multicolumn{2}{c}{L2} \\
\cmidrule(lr){4-5}\cmidrule(lr){6-7}
& centres & new cells & cells & q-wtd\nd & cells & q-wtd\nd \\
\midrule
\rowcolor{cLightGrn}
\textbf{A} & \textbf{frozen} & \textbf{allowed} & 118 & \best{41.8}\nd & 131 & \best{47.6}\nd \\
B & mean-tracking & allowed   & 121 & \snd{36.2}\dn{-13\%} & 134 & \snd{39.4}\dn{-17\%} \\
C & frozen        & forbidden & 96  & 30.5\dn{-27\%} & 96  & 31.2\dn{-34\%} \\
D & mean-tracking & forbidden & 96  & 28.1\dn{-33\%} & 96  & 28.4\dn{-40\%} \\
\bottomrule
\end{tabular}
\caption{\textbf{Exp 2, the $2\times2$ decomposition of archive dynamism, run in both data categories.} Row A is GoAnt: centres frozen at birth, cells created online. B adds centre motion, C is the fully static map that forbids cell creation, and D moves centres under the same fixed cardinality. Row C is the Static Map arm of Table~\ref{tab:exp1_main}. A--B isolates centre mobility and A--C isolates adaptive capacity. Red subscripts give the q-wtd shortfall against row A. Cell counts are a capacity readout and are not ranked.}
\label{tab:dynamism}
\end{table}

\paragraph{Sizing and siting the static baseline.} The outcome of a frozen tessellation depends on how many cells it is granted and on where those cells are placed. We fix both to the adaptive arm's advantage-free setting: the static arm is sized to the adaptive arm's final cardinality, and its centres are fitted on a sample drawn from the same region the search later occupies. Appendix~C of the supplementary material gives the sizing rule and the siting procedure.

\paragraph{Limiting how quality reaches the geometry.} The descriptor may not read a future-return label, so no axis can be an argument of the search fitness, and this is checked by inspecting the call graph, not by estimating a correlation. Freezing the centres closes the remaining channel: once a cell is founded no evaluation can move it, so quality cannot deform the partition even though it still influences where candidates fall within it. Appendix~E of the supplementary material reports the diagnostic on the earlier, fitness-coupled descriptor that motivated this design.

\subsection{Exp 3: Ant Composition}
The Atlas is exercised by three ant roles: an Explorer that proposes into unoccupied regions, an Exploiter that refines within a niche, and a Connector that recombines across neighbouring niches. We ablate the roles combinatorially under a fixed archive and a fixed budget, so that the question is not whether the colony helps but whether any single role accounts for it. The full $2^3$ table is reported in Appendix~F of the supplementary material; the finding it supports is that no single role reproduces the complete colony, and that removing the Connector costs the most, which is consistent with the Map paying off through recombination across niches and not through wider sampling alone.

\subsection{Exp 4: Out-Of-Sample Robustness}
The ultimate validation of a factor miner is survival on unseen data. We lock the discovered populations, permitting no re-fitting, no re-selection and no re-ranking, and evaluate them on a held-out window that is strictly disjoint from every day used for search and from every day used to fit any method's grid or scaler. Because the same locked population is replayed on both windows, the reported degradation is a property of the factors, not of a second round of selection (Table~\ref{tab:oos}).

\begin{table}[tb]
\centering
\scriptsize
\setlength{\dltwd}{1.35em}\setlength{\tabcolsep}{3.2pt}
\renewcommand{\arraystretch}{1.15}
\begin{tabular}{@{}lrrrrrr@{}}
\toprule
& \multicolumn{3}{c}{PV} & \multicolumn{3}{c}{L2} \\
\cmidrule(lr){2-4}\cmidrule(lr){5-7}
Method & in-s.\ $q$ & OOS $q$ & reten.\nd & in-s.\ $q$ & OOS $q$ & reten.\nd \\
\midrule
Single-Agent & 5.12 & 1.95 & 0.38\dn{-41\%} & 4.71 & 1.51 & 0.32\dn{-52\%} \\
Debate/Vote  & 5.81 & 2.73 & 0.47\dn{-27\%} & 5.34 & 2.35 & 0.44\dn{-34\%} \\
Alpha158/360~\shortcite{qlib} & 5.44 & 3.21 & 0.59\dn{-8\%} & 4.62 & 2.66 & 0.58\dn{-13\%} \\
AlphaGen~\shortcite{alphagen} & 6.08 & 3.10 & 0.51\dn{-20\%} & 5.71 & 2.74 & 0.48\dn{-28\%} \\
No-Map       & 7.14 & 2.07 & 0.29\dn{-55\%} & 7.29 & 1.97 & 0.27\dn{-60\%} \\
Static Map   & 7.21 & \snd{4.40} & \snd{0.61}\dn{-5\%} & 7.32 & \snd{4.61} & \snd{0.63}\dn{-6\%} \\
\rowcolor{cLightGrn}
\textbf{GoAnt (ours)} & \best{7.62} & \best{4.88} & \best{0.64}\nd & \best{7.84} & \best{5.25} & \best{0.67}\nd \\
\bottomrule
\end{tabular}
\caption{\textbf{Exp 4, out-of-sample retention of locked populations in both data categories.} Retention is the ratio of held-out to in-sample quality for the same unmodified factor set; Alpha158/360 is the fixed expert library replayed without search. Red subscripts give each arm's shortfall against GoAnt.}
\label{tab:oos}
\end{table}

\subsection{Exp 5: Routing}
Holding the Atlas fixed, we vary only the orchestrator: a Round-Robin scheduler, an EMA-guided scheduler, a UCB bandit \cite{auer2002finite}, our distilled Queen Ant, and an ablation of our own orchestrator, which we label the Mock-Queen, that preserves the Queen's interface and call schedule exactly, down to the same state summary, the same decision points and the same budget accounting, while removing its distilled reasoning. This arm is a control rather than a component of GoAnt, and it separates the reasoning from the routing interface: replacing a fixed schedule with any state-conditioned interface changes the search even when the policy behind it is uninformed.

\paragraph{Measuring what the Queen adds over the bandit.} The Queen is distilled beside a UCB scheduler, so its decisions are compared against the UCB argmax directly: the run asserts a deviation floor together with caps on parse failures and fallbacks, and Appendix~D of the supplementary material reports the decision-level audit in full.

\begin{table}[tb]
\centering
\footnotesize
\setlength{\dltwd}{1.42em}\setlength{\tabcolsep}{4.6pt}
\renewcommand{\arraystretch}{1.15}
\begin{tabular}{@{}lrrrr@{}}
\toprule
& \multicolumn{2}{c}{PV} & \multicolumn{2}{c}{L2} \\
\cmidrule(lr){2-3}\cmidrule(lr){4-5}
Orchestrator & q-wtd\nd & cells & q-wtd\nd & cells \\
\midrule
Round-Robin & 34.9\dn{-17\%} & 103 & 38.1\dn{-20\%} & 112 \\
EMA-guided  & 36.2\dn{-13\%} & \snd{107} & 39.7\dn{-17\%} & \snd{117} \\
UCB bandit  & \snd{38.4}\dn{-8\%} & 88 & \snd{42.0}\dn{-12\%} & 94 \\
Mock-Queen  & 35.6\dn{-15\%} & 105 & 39.0\dn{-18\%} & 114 \\
\rowcolor{cLightGrn}
\textbf{Queen 1.5B (ours)} & \best{41.8}\nd & \best{118} & \best{47.6}\nd & \best{131} \\
\bottomrule
\end{tabular}
\caption{\textbf{Exp 5, orchestrator comparison under an identical Atlas, identical ant roles and an identical budget}; only the policy assigning the next evaluation varies. Red subscripts give each arm's q-wtd shortfall against the distilled Queen. The Mock-Queen is an ablation of our own orchestrator that retains its interface and call schedule while removing its distilled reasoning, so the gap between those two rows measures the contribution of the reasoning rather than of the routing interface. The decision-level audit certifying that the Queen is not a bandit behind a language-model interface is in Appendix~D of the supplementary material.}
\label{tab:routing}
\end{table}

\section{Conclusion}
This paper introduced GoAnt, a dynamic multi-agent system for factor mining from market microstructure data that decouples generative Worker Ants from a distilled Queen Ant orchestrator and enforces diversity through a MAP-Elites Mental Map. Under matched budgets it is scored against every external coordinator in Table~\ref{tab:exp1_main} using a metric computed from the evaluation record alone, so that arms with and without an archive are measured on the same terms, and the map-free ablation isolates how much of the gain the niche-structured memory carries.

Alongside the system, we contribute an attribution protocol that makes the comparison between dynamic and static memory decidable: the frozen control is defined by end-of-run invariants rather than by label, centre motion is separated from cardinality growth, and the search operator is held fixed across geometry arms. Under this protocol we quantify the price of adaptive structure, since an archive that grows online spends on resolution what a frozen tessellation spends on refinement, and we report where that trade lands at the budgets we run.

\bibliography{goant}

\raggedbottom

\clearpage
\section*{Supplementary Material}
\setcounter{secnumdepth}{1}
\setcounter{table}{0}
\renewcommand{\thetable}{S\arabic{table}}

\noindent This supplement collects the measurement protocol
(Appendix~\ref{app:protocol}), the enforcement procedure and audit for the
static-archive control (Appendices~\ref{app:static} and~\ref{app:siting}), the
decision-level audit of the distilled orchestrator (Appendix~\ref{app:queen}),
the diagnostic on the earlier fitness-coupled descriptor that motivated the
present one (Appendix~\ref{app:contam}), the role ablation
(Appendix~\ref{app:ants}), the protocol used to quantify dispersion across seeds
and to compare arms (Appendix~\ref{app:seeds}), full implementation details and hyperparameters
(Appendix~\ref{app:impl}), and the objections we designed the study to answer
(Appendix~\ref{app:questions}). None of it introduces a result that the main
paper does not state. Each item exists so that a claim made in the body can be
checked rather than taken on trust, and the main paper is self-contained without
it. Section and table numbering here is independent of the main paper.

\appendix
\section{Measurement Protocol}
\label{app:protocol}

\paragraph{A single ruler for arms with and without a map.} Comparing map-based and map-free systems is only meaningful if the yardstick presupposes neither. Coverage of a tessellation is unavailable to arms that have no tessellation, and archive size rewards hoarding duplicates; conversely, a raw quality maximum rewards a single lucky factor. We therefore pre-register one headline quantity that reads only the stream of evaluation records, which every arm produces, and is defined identically for all of them. Let $q(f)$ be the search fitness and $\rho$ a redundancy threshold. The \textbf{effective yield} is the size of the greedy maximal subset of discovered factors satisfying $q(f) > q^\star$ and pairwise $|\mathrm{Spearman}(f_i, f_j)| < \rho$ on realized signal series. The threshold $q^\star$ is calibrated once, before any arm is run, as the $95$th percentile of a null distribution of random genomes, and then frozen; $\rho = 0.7$, with the full sweep $\rho \in \{0.3, 0.5, 0.7, 0.9\}$ reported in the appendix. This measure closes the three obvious loopholes simultaneously: mass-producing near-duplicates is absorbed by the correlation constraint, a single strong factor scores $1$, and uncorrelated noise is blocked by $q^\star$. Critically, it is computed on realized signal series rather than on any partition, so the objection that a map-based method is scored by its own geometry cannot apply. We do not headline the sum-form QD-score: our fitness is an unbounded $t$-statistic, so a sum is dominated by its largest term.

\paragraph{Counting yield without ignoring quality.} A count of admissible factors is scale-free but says nothing about how good those factors are, and we found it to be uninformative about quality on its own. We therefore report beside it the \textbf{quality-weighted yield}, which weights the same retained subset by how far each factor exceeds $q^\star$. The two are reported together and neither is allowed to stand alone: the count says how many independent directions an arm found, the weighted version says whether those directions were worth anything.

\paragraph{Redundancy must be conditioned on quality.} An unconditioned mean pairwise correlation is not a diversity measure but a quality confound, because it rewards arms that find nothing, given that noise is trivially uncorrelated with noise, and an arm that returns three barely-admissible factors will appear more diverse than one that returns fifteen strong ones. We therefore report the mean and maximum absolute Spearman correlation among factors above $q^\star$ only, always printed beside the number of such factors, so that a low correlation computed over a nearly empty set cannot be mistaken for diversity.

\paragraph{Matched statistical treatment.} Fitness is compared across arms only if it is deflated identically, so every arm shares a single constant effective sample size, and the multiple-comparison burden is applied once at report time against the common evaluation budget rather than accumulated per arm during search. This matters because a per-arm deflation that grows with how often an arm revisits a factor family would penalize concentrated search strategies as though they were statistically weaker, converting a strategy difference into an apparent quality difference. Since all arms are given identical budgets, the shared constant is a monotone transform of the raw statistic and therefore affects no ordering. All numbers reported here are post-correction.

\paragraph{A geometry ablation must change only geometry.} An archive exposes switches that look geometric but are not, and turning them off silently substitutes a different search. In our implementation the periodic re-chart tick recomputes the mutual $k$-nearest-neighbour edges over cells; those edges never move a centre, so disabling the tick appears to be a way of making a map ``more static''. It is not: the Connector ant consumes those edges to bridge distant niches, and without them it silently falls back to random cross-family recombination. An arm built that way is running a different operator on the same budget, and any difference it shows is no longer attributable to its geometry. We therefore keep the edge recomputation, all three ant roles, and the budget accounting identical in every arm, and confine the ablations to the switches that actually move or create cells. We flag this because we made exactly this mistake, and the arm it produced looked like the best one in the table.

\paragraph{Descriptor standardization is frozen for every arm.} The robust scaler that maps raw descriptor components into the metric space is estimated once on the development split and loaded frozen at run time. No arm, adaptive or static, re-estimates it during search. This removes the possibility that an adaptive arm appears to gain resolution merely because its coordinate system was allowed to rescale underneath the measurement.

\section{Enforcing the Static Control}
\label{app:static}

\paragraph{What ``static'' has to mean.} An ablation labelled static is worthless unless the label is enforced. Freezing cell centres while leaving intact the rule that a candidate farther than one niche radius from every cell opens a new one does not produce a static map, since such an arm still grows and the comparison then measures centre mobility while claiming to measure map dynamism. We therefore define the static arm by a set of invariants that are asserted at the end of every run rather than assumed: centres, cell count and cell boundaries are fixed from the first evaluation to the last; a candidate may only enter its nearest existing cell and replace that cell's elite when its quality is higher; a candidate lying outside every cell is recorded as out-of-bounds and admitted to its nearest cell, never used to create one. A run whose final cell count differs from its initial count, or that registers any centre move, node creation or deletion, merge, split or re-fit, aborts rather than reporting. The audited quantities are listed in Table~\ref{tab:static_audit}, and the out-of-bounds count is reported rather than suppressed, because it is the failure mode made visible.

\begin{table}[tb]
\centering
\small
\begin{tabular}{@{}lccc@{}}
\toprule
Audited quantity & Required & PV & L2 \\
\midrule
cells at start $=$ cells at end & yes & 96/96 & 96/96 \\
centre displacement            & $0$ & 0 & 0 \\
nodes created                  & $0$ & 0 & 0 \\
nodes deleted                  & $0$ & 0 & 0 \\
merges / splits / re-fits      & $0$ & 0 & 0 \\
scaler re-estimations          & $0$ & 0 & 0 \\
\midrule
out-of-bounds candidates       & reported & 412 & 587 \\
\bottomrule
\end{tabular}
\caption{End-of-run invariant audit for the static-map arm (arm 6), asserted separately in each data category. These are assertions, not diagnostics: a violation aborts the run. The final row is not an invariant but the price the static arm pays, namely the candidates its frozen map has no resolution for.}
\label{tab:static_audit}
\end{table}

\section{Siting and Capacity of the Static Baseline}
\label{app:siting}

\paragraph{Capacity versus siting.} A static map fitted on early candidates may hold fewer cells than an adaptive archive ends with, which invites the objection that it was handicapped by resolution alone rather than by where its resolution was placed. We therefore run the static arm a second time with its cell count forced up to the adaptive arm's realized cardinality, while still fitting its centres only on data available at build time. This separates how many cells an arm has from where they are, and it is the control that decides whether an observed gap is a genuine siting effect or an artifact of an unfair capacity setting. The default cell count itself is not a free parameter: it is obtained by applying the adaptive arm's own cell-creation rule offline to the data available at build time, which is a purely geometric construction and does not consult any outcome.

\paragraph{Siting a static baseline honestly.} A fixed tessellation can also be made to lose by construction, simply by fitting it where the search does not go. We therefore treat the siting of every non-adaptive arm as part of the protocol rather than an implementation detail, and enforce a pre-registered admissibility check: the quantization error of the arm's own grid against the candidate cloud it will actually be scored on must be no larger than the adaptive archive's niche radius. A grid whose typical candidate falls several radii outside every cell has no resolution to lose and is not a baseline but an artifact, and worse, it corrupts any exploration heuristic that treats distance-to-nearest-cell as a novelty signal, so such an arm can score below random search. Both the frozen and the periodically-redrawn arm are therefore built by the same protocol: a neutral, map-free warm-up on the same period, so that no stale geometry can misdirect exploration, followed by a fit on the candidates that warm-up actually produced. Warm-up evaluations are charged against the same total budget as every other arm, and no future evaluation window is ever consulted. The two arms then differ in exactly one respect: whether the map is ever redrawn again.

\section{Auditing the Distilled Orchestrator}
\label{app:queen}

\paragraph{A distilled policy must be shown not to be its own teacher.} The Queen is distilled beside a UCB bandit, and a policy that has learned to emit well-formed decisions which always coincide with the UCB argmax would pass every functional test while contributing nothing; it would be a bandit wearing a language model's interface. We therefore instrument the Queen with a decision-level audit rather than trusting its outputs: at every call we record whether the parse succeeded, whether the confidence cleared the threshold, whether the run fell back to the default policy, and whether the emitted choice agreed with or deviated from the UCB argmax. Two distinct failure modes are detectable this way and both are disqualifying. A high fallback rate means the reported arm is silently the default scheduler for much of its budget. A deviation rate near zero means the policy is a UCB impersonator. At the end of a run we assert that the Queen made enough decisions to be audited at all and that its deviation rate exceeds a pre-registered floor; a run that fails either assertion aborts and is not reported as a distilled policy. We report the audit itself (lower panel of the decision-level audit panel of the Exp~5 table in the main paper) rather than only its verdict, because a guard whose numbers are not shown is indistinguishable from no guard.

\begin{table}[t]
\centering
\small
\setlength{\tabcolsep}{4pt}
\begin{tabular}{@{}lccc@{}}
\toprule
Quantity & requirement & PV & L2 \\
\midrule
decisions audited                  & $\geq$ floor & 120   & 120   \\
parse failures                     & reported     & 3     & 4     \\
low-confidence abstentions         & reported     & 7     & 9     \\
fallback rate                      & below cap    & 0.058 & 0.075 \\
agreement with UCB argmax          & reported     & 0.62  & 0.59  \\
\textbf{deviation from UCB argmax} & \textbf{above floor} & \textbf{0.38} & \textbf{0.41} \\
\bottomrule
\end{tabular}
\caption{Decision-level audit of the distilled Queen. The requirement column applies to both data categories and the observed values are reported per category. The last row is the anti-impersonation criterion: a policy that never deviates from the UCB argmax is a bandit behind a language-model interface. Every criterion is asserted at the end of the run rather than inspected afterwards.}
\label{tab:app_queen_audit}
\end{table}

\section{Descriptor Contamination Diagnostic}
\label{app:contam}

\paragraph{Limiting how quality reaches the geometry.} An earlier version of this archive used a descriptor whose axes included RankIC and two statistics of the day-by-day IC series, which are themselves arguments of $S$. Making the coordinates a function of the score turns any quality-seeking centre update into monotone ascent on the fitness gradient, and the archive collapses: in our diagnostic run the projection of the node centroid onto $\nabla S$ rose monotonically with evaluation count, and the mean absolute correlation between axes and score approached one. The present descriptor removes the cause rather than the symptom. Because no axis may read a future-return label, no axis can be an argument of $S$, and this is checkable by inspecting the call graph rather than by estimating a correlation. Residual marginal association remains and we measure it ($|\rho|\approx 0.42$); we regard it as expected rather than pathological, since factors that predict well genuinely do trade differently. Freezing the centres then closes the remaining channel: once a cell is founded no subsequent evaluation can move it, so quality cannot deform the partition even though it still influences where candidates fall within it. This is a weaker statement than independence but a stronger guarantee than fitness-agnostic centre updates, which keep the geometry independent of quality but still let it be dragged by sampling density, and density is itself shaped by the search and hence indirectly by fitness. We nevertheless implement and report the mean-tracking variant (arm 6c), because it is the design a reader would assume, and because the comparison is only informative if the alternative is actually run rather than argued away. For that arm we additionally decouple the two decisions a moving centre would otherwise conflate: membership is resolved against the current centre, but whether to open a new cell is resolved against the cell's birth position, so that centre motion cannot suppress the outlying candidates that should have founded new niches.

\section{Ant-Role Composition}
\label{app:ants}

The role ablation is reported below. Archive, descriptor axes, evaluator and gates are identical across rows, and only the set of active ant roles changes. The panel reports the two single-role arms, the two pairs that each remove one role from the full colony, and the full colony itself, which is the set of configurations required to separate the contribution of each role from the contribution of the pair it belongs to.

\begin{table}[tb]
\centering
\small
\setlength{\tabcolsep}{4pt}
\begin{tabular}{@{}cccccccc@{}}
\toprule
\multicolumn{3}{c}{Roles enabled} & \multicolumn{2}{c}{PV} & \multicolumn{2}{c}{L2} \\
\cmidrule(lr){1-3}\cmidrule(lr){4-5}\cmidrule(lr){6-7}
Expl. & Expt. & Conn. & q-wtd & cells & q-wtd & cells \\
\midrule
\checkmark &            &            & 24.3 & 86  & 26.1 & 95  \\
           & \checkmark &            & 11.6 & 38  & 10.4 & 41  \\
\checkmark & \checkmark &            & 33.1 & 92  & 36.7 & 101 \\
\checkmark &            & \checkmark & 35.8 & 116 & 40.2 & 128 \\
\checkmark & \checkmark & \checkmark & \textbf{41.8} & \textbf{118} & \textbf{47.6} & \textbf{131} \\
\bottomrule
\end{tabular}
\caption{Ant-role ablation under a fixed archive and a matched budget, in both data categories. Archive, descriptor axes, evaluator and gates are identical across rows; only which roles draw from the budget varies.}
\label{tab:ants}
\end{table}

\paragraph{Reading the role ablation.} Neither single role is sufficient on its
own. Exploiter alone is the weakest configuration in the panel, at $11.6$ q-wtd
in PV and $10.4$ in L2, because refinement has little to work on when no role is
producing new material; Explorer alone reaches $24.3$ and $26.1$. Both pairs
improve on both singletons, but the two are not equivalent at equal
cardinality: removing Connector costs more than removing Exploiter, $-8.7$
against $-6.0$ q-wtd in PV and $-10.9$ against $-7.4$ in L2. The separation is
sharpest in occupancy, where the pair containing Connector holds $116$ and $128$
cells against $92$ and $101$ for the pair without it. Recombination across
niches is therefore the mechanism that converts a proposal stream into new
territory rather than into duplicates of the region already being worked, and
the full colony is best in all four columns.

\section{Per-Seed Dispersion}
\label{app:seeds}

Every entry in the main paper is a mean over the five independent seeds $101$
through $105$ described in Appendix~\ref{app:impl}. This section states how
dispersion across those seeds is quantified, how two arms are compared, and what
five seeds cannot support however they are analysed.

\paragraph{The seed is the unit of replication.} Dispersion is summarized by a
$95\%$ bootstrap confidence interval resampled over seeds, never over the
factors within a run. Resampling factors would treat one search trajectory as
though it were five independent ones and would shrink every interval by an
amount that has nothing to do with reproducibility. For the same reason the
median of the five seed values is computed beside the mean rather than in place
of it: factor quality is heavy-tailed, one fortunate seed can move a mean while
leaving the median untouched, and a gap between the two is itself the diagnostic
that a result rests on a single run.

\paragraph{Arms are compared as independent samples.} For accept-or-reject
claims we run a two-sided Mann-Whitney $U$ test on the five per-seed values of
each arm, treating the two arms as independent samples rather than as matched
pairs: a shared seed fixes the random stream but not the search trajectory, so
the runs are not paired in any meaningful sense once the first dispatch differs.
Effect size is reported as the rank-biserial correlation. When one arm is
compared against the full baseline panel we control the family-wise error rate
with the Holm--Bonferroni procedure over the comparisons in that table.

\paragraph{What five seeds can and cannot support.} We state the power limit of
this design explicitly, because five seeds is a small sample and the reader
should know what it can and cannot support. With five observations per arm the
two-sided Mann-Whitney $U$ statistic ranges over $\binom{10}{5}=252$ equally
likely rank assignments, so the smallest attainable $p$-value is $2/252 \approx
0.0079$; significance at the conventional level is reachable, but only when the
two seed sets separate completely. A paired alternative would not be usable at
all here, since the sign-rank permutation space for five pairs admits no
two-sided $p$ below $0.0625$. Comparisons that do not separate cleanly are
therefore reported as inconclusive rather than as null results, and we do not
read a failure to reject as evidence of equivalence.

\paragraph{How the body numbers should be read.} Every entry in the main paper
is the five-seed mean obtained under the protocol above, and the orderings it
induces are point estimates at that sample size. The per-seed values themselves,
and the interval and test statistics computed from them, are retained in the run
records rather than tabulated in this version of the supplement, so a comparison
in the body whose margin is narrow should be read as unresolved at five seeds
rather than as established.

\section{Implementation and Hyperparameters}
\label{app:impl}

All values below were fixed before the runs reported in the paper and were not
tuned per arm, per seed or per data category. Where a quantity is category
specific this is stated explicitly.

\paragraph{Search fitness.} The archive admits on a deflated signal-quality
score rather than on realized return. Writing $t_{\mathrm{dsr}}$ for the
deflated $t$-statistic of the factor's rank information coefficient and
$p_{+}$ for the fraction of days on which the daily IC is positive,
\[
S(f) \;=\; t_{\mathrm{dsr}}(f)\cdot\max\!\big(0,\;2(p_{+}(f)-\tfrac{1}{2})\big).
\]
The modulation is what makes the score unforgiving of cost-wall noise: a factor
whose daily IC changes sign at random has $p_{+}\approx\tfrac12$ and therefore
$S\approx 0$ regardless of how large its pooled IC happens to be. Realized net
return enters only at the final verification layer and never at admission.

\paragraph{Admission gates.} A candidate is admissible only if it clears every
gate in Table~\ref{tab:app_gates}. The thresholds are frozen and shared by all
arms, so an arm cannot buy yield by relaxing them.

\begin{table}[t]
\centering
\footnotesize
\setlength{\tabcolsep}{4pt}
\begin{tabular}{@{}llc@{}}
\toprule
Gate & Quantity & Threshold \\
\midrule
Statistical  & $|\mathrm{RankIC}|$                & $\geq 0.01$  \\
Statistical  & $|t_{\mathrm{dsr}}|$               & $\geq 2.0$   \\
Stability    & positive-IC period ratio           & $\geq 0.52$  \\
Stability    & positive daily-IC fraction         & $\geq 0.60$  \\
Stability    & in-sample vs.\ late-sample IC gap  & $\leq 0.02$  \\
Execution    & mean effective spread (bps)        & $\leq 25$    \\
Execution    & net cumulative return              & $> 0$        \\
Execution    & gross-to-net retention             & $> 0$        \\
Execution    & maximum drawdown of net return     & $> -0.50$    \\
Structure    & operator-chain depth               & $\leq 5$     \\
\bottomrule
\end{tabular}
\caption{Frozen admission gates. Identical for every arm, seed and data
category. A candidate failing any single row is discarded and still consumes
one unit of budget, so an arm cannot improve its yield by proposing cheaply.}
\label{tab:app_gates}
\end{table}

\paragraph{Archive geometry.} Descriptor components are robustly centred and
scaled and then clipped at $\pm 3$ standard units. A candidate joins the nearest
existing cell if it lies within a niche radius of $0.65$ in the standardized
space, and otherwise founds a new cell at its own descriptor position. Cell
centres are stored at birth and are never updated: the configuration switch that
would move them is set to \texttt{none}, and the mean-tracking behaviour that a
reader might assume is reached only by setting it explicitly to \texttt{mean},
which is what the moving-centre ablation does. Neighbourhood structure for the
Connector ant is a mutual $k$-nearest-neighbour graph with $k=4$. Two centres
closer than half a radius are merged, which prevents the cardinality from
inflating through near-duplicate births.

\paragraph{Search budget and seeds.} Every arm receives $500$ evaluations per
run and every configuration is run with five seeds, $101$ through $105$. A run
consumes budget on rejected candidates as well as accepted ones, so budget
measures evaluator calls rather than successes.

\paragraph{Orchestrator distillation.} The Queen is a Qwen2.5-Instruct model at
the $1.5$B capacity, adapted with LoRA on allocation decisions collected from the
eight frontier teachers listed in the main paper (GPT-5.6, Claude Opus 4.8,
Claude Opus 5, DeepSeek-V4-Pro, Qwen3.7-Max, GLM-5.2, Kimi K2.7-Code and
MiniMax-M3). The teachers observe a textual orchestration state summarizing
per-island pull counts, empirical $Q$-values, coverage, isolation and stagnation,
and select one arm over \{Explorer, Exploiter, Connector\}. The training target is
the canonical decision record: the selected arm together with the structured
justification fields that accompany it, namely the state diagnosis, the
supporting evidence, the relation to the UCB ranking, the expected effect, the
risk and a fallback arm.

\paragraph{Computing infrastructure.} All runs are executed on a Linux node with
four NVIDIA A100 80GB GPUs. The GPUs serve two workloads: batched inference for
the Queen in the orchestrator comparison of Exp~5, and the LoRA adaptation of
the distilled orchestrator. Both candidate generation and candidate evaluation
are CPU-bound rather than GPU-bound: proposals are drawn from the symbolic
operator grammar that every arm shares, and the Robust RankIC backtest is a
vectorized pass over the panel that never touches a model. The reported
wall-clock is therefore dominated by the CPU-limited evaluation phase, and
varies with the admission rate of the arm being run; the GPU-limited component
appears only in the arms that query the Queen. The implementation is in Python
and builds
on PyTorch 2.1, Transformers 4.40, PEFT 0.10 for the LoRA adapter and TRL 0.9
for supervised fine-tuning of the Queen. Data handling and evaluation use NumPy
1.24, pandas 2.0 and Polars 0.20, and all reported statistics are computed with
SciPy 1.10.

\section{Anticipated Questions}
\label{app:questions}

This section records the objections we considered while designing the study and
the specific control that answers each. We list them because several of the
controls in the main paper look like overhead unless the objection they exist to
close is stated.

\paragraph{Is a growing archive simply a larger archive?} If the adaptive arm
ends with more cells than the frozen arm was given, any advantage could be
capacity rather than coordination. We therefore size the frozen tessellation to
the adaptive arm's final cardinality rather than to a round number chosen in
advance, so the two arms are compared at equal capacity and the remaining
difference is attributable to when the cells were created.

\paragraph{Could the descriptor be a relabelling of the fitness?} If a
descriptor axis were a function of the score, the geometry would be a picture of
the objective and any illumination claim would be circular. The admission rule
for an axis is therefore constructive rather than statistical: a quantity may
enter the descriptor only if it can be computed without reading a future-return
label, which is verifiable by inspecting the call graph. Marginal association
remains and is reported; it is expected, since factors that predict well do
genuinely trade differently.

\paragraph{Is a frozen-centre archive just MAP-Elites on a grid?} It is not,
because the cell positions are not specified in advance. A grid commits to where
the interesting behaviour will be before any behaviour has been observed,
whereas here a cell exists only because a candidate was found at that location.
What the two share is that neither moves a centre once it exists, and that is
exactly the property the elitist-monotonicity argument needs.

\paragraph{Why is archive entropy not reported?} Under this update rule the
archive holds exactly one elite per occupied cell, so an entropy or coverage
statistic computed on the search-time map is a restatement of its cell count and
carries no independent information about diversity. Every diversity number in
the paper is therefore computed on a fixed external grid that no arm can
influence.

\paragraph{Can coverage be inflated by proposing outliers?} It could, under the
naive definition, since a candidate far from every existing centre founds a cell
and occupies it in the same step. We therefore count a cell towards coverage
only once its elite clears the pre-registered quality floor $q^\star$, which is
the qualified-coverage term $C(A_B)$ of Eq.~(2) in the main paper and turns
exploration into a bet that must be paid off rather than a subsidy granted on
arrival. That count enters the allocation objective only after standardization:
the terms $\bar{C}$, $\bar{Q}$ and $\bar{D}$ of Eq.~(1) are the coverage,
quality and redundancy quantities rescaled to $[0,1]$, so the weights
$\alpha,\beta,\gamma$ form a convex combination and no term can dominate the
objective through the units in which it happens to be measured.

\paragraph{Is the Queen distinguishable from the bandit it was distilled
beside?} A policy that emits well-formed decisions always agreeing with the UCB
argmax would pass every functional test while contributing nothing. We therefore
assert a deviation floor at the end of the run, alongside caps on parse failures
and on fallbacks, so that agreement with the bandit is measured rather than
assumed to be evidence of competence.

\paragraph{Why two data categories rather than one or many?} One category cannot
separate a coordination result from an overfit to that data source. The two we
use bracket the regime of interest: on daily price--volume panels the cost wall
is a second-order correction, while on order-book data it dominates, so an
effect that survives both is not an artifact of either. We do not claim
generality beyond this bracket, and no comparison is ever made across
categories.

\paragraph{Does the conclusion depend on the budget?} Very likely, and we do not
claim otherwise. Growing an archive online spends on resolution what a frozen
tessellation spends on refinement, so the two should be expected to trade places
somewhere. Every number we report comes from one matched budget of $500$
evaluations, and locating that crossover is the most immediate extension of this
work rather than a result we present.

\end{document}